\documentclass[conference,10pt]{IEEEtran}
\IEEEoverridecommandlockouts
\usepackage{cite}
\usepackage{amsmath,amssymb,amsfonts}
\usepackage{graphicx}
\usepackage{textcomp}
\usepackage{xcolor}
\def\BibTeX{{\rm B\kern-.05em{\sc i\kern-.025em b}\kern-.08em
    T\kern-.1667em\lower.7ex\hbox{E}\kern-.125emX}}
\def\noIndentSpace{4pt}
\usepackage{comment}
\usepackage{subcaption}
\usepackage{graphicx}
\usepackage{algorithm, algcompatible}
\usepackage{algpseudocode}
\usepackage[colorlinks=true,linkcolor=magenta,citecolor=cyan]{hyperref}%
\usepackage{booktabs}
\usepackage{float}
\usepackage{enumitem}
\usepackage{balance}
\usepackage[labelsep=period,labelfont=bf]{caption}
\usepackage{subcaption}
\usepackage{tcolorbox,tikz}
\tcbuselibrary{skins}
\tcbuselibrary{breakable}
\usepackage{threeparttable} 

\newtcolorbox{mybox}[3][]
{
  breakable, 
  enhanced,
  colback  = #2!10, 
  colframe= #2!50,
  boxsep=-0.5mm,
  borderline west={0.1mm}{0.05mm}{#3!5}, 
  #1,
}    

\begin{document}

\title{Cross-Platform Scaling of Vision-Language-Action Models from Edge to Cloud GPUs \\
}
\author{
    Amir Taherin\IEEEauthorrefmark{1},
    Juyi Lin\IEEEauthorrefmark{1},
    Arash Akbari\IEEEauthorrefmark{1},
    Arman Akbari\IEEEauthorrefmark{1},
    Pu Zhao\IEEEauthorrefmark{1},
    Weiwei Chen\IEEEauthorrefmark{2},
    David Kaeli\IEEEauthorrefmark{1},
    Yanzhi Wang\IEEEauthorrefmark{1} \\
    \IEEEauthorblockA{\IEEEauthorrefmark{1}Department of Electrical and Computer Engineering, Northeastern University, Boston, MA \\
    Email: \{taherin.a, lin.juy, akbari.ara, akbari.ar, p.zhao, d.kaeli, yanz.wang\}@northeastern.edu}
    \IEEEauthorblockA{\IEEEauthorrefmark{2}EmbodyX Inc, Belmont, CA, Email: weiwei.chen@embodyx.io}
}

\maketitle

\begin{abstract} 
Vision-Language-Action (VLA) models have emerged as powerful generalist policies for robotic control, yet their performance scaling across model architectures and hardware platforms, as well as their associated power budgets, remain poorly understood. This work presents an evaluation of five representative VLA models---spanning state-of-the-art baselines and two newly proposed architectures---targeting edge and datacenter GPU platforms. Using the LIBERO benchmark, we measure accuracy alongside system-level metrics, including latency, throughput, and peak memory usage, under varying edge power constraints and high-performance datacenter GPU configurations. Our results identify distinct scaling trends: (1) architectural choices, such as action tokenization and model backbone size, strongly influence throughput and memory footprint; (2) power-constrained edge devices exhibit non-linear performance degradation, with some configurations matching or exceeding older datacenter GPUs; and (3) high-throughput variants can be achieved without significant accuracy loss. These findings provide actionable insights when selecting and optimizing VLAs across a range of deployment constraints.  Our work challenges current assumptions about the superiority of datacenter hardware for robotic inference.
\end{abstract}


\section{Introduction}
\label{sec:introduction}
Vision-Language-Action (VLA) models have recently emerged as a powerful paradigm for robotic control, enabling systems to perceive the environment, reason over task instructions, and generate executable actions directly from vision and language inputs~\cite{o2024open, kim2024openvla, lin2025vote, kim2025fine, qu2025spatialvla, zitkovich2023rt}. By integrating vision and language backbones with specialized action heads~\cite{lin2025vote}, VLAs have demonstrated impressive generalization across diverse robotic tasks, from tabletop manipulation to long-horizon planning~\cite{o2024open, lin2025vote}. These advances open the door to deployment of a single, generalist policy across a variety of robots and environments, reducing the need for custom-made task-specific models~\cite{o2024open,kim2024openvla}. As robotic systems increasingly transition from controlled laboratory settings to real-world applications, the ability to run such models efficiently across a variety of hardware platforms—from low-power edge devices to high-performance datacenter GPUs—has become a critical requirement.

Despite rapid algorithmic progress, little is known about how VLA performance scales across different \textit{model architectures}, \textit{hardware classes}, and \textit{power budgets}. Most prior studies focused on improving model accuracy or adapting vision-language architectures for action generation, but they typically perform evaluation on a single hardware platform under fixed resource settings~\cite{kim2024openvla,kim2025fine}. As a result, there is a limited understanding of the trade-offs between accuracy, latency, throughput, and memory usage when deploying VLAs across the full spectrum of computing platforms—from power-constrained edge devices to high-performance datacenter GPUs. This gap limits our ability to make informed deployment decisions, and optimize models for specific environments.

Understanding these scaling trends is essential for designing and deploying VLAs in real-world robotic systems, where hardware resources, latency requirements, and energy budgets vary widely. In embedded platforms such as service robots, mobile manipulators and autonomous drones, computational and power constraints necessitate striking a careful balance in terms of model size, throughput, and accuracy to ensure responsive and reliable operation~\cite{edgeAI_robotics}. In cloud-hosted or datacenter settings, maximizing throughput, while containing memory usage, can directly impact operational costs and scalability. Without a systematic view of how VLA architectures behave across this edge–cloud spectrum, engineers risk over-provisioning hardware and underutilizing available resources, making suboptimal trade-offs between performance and efficiency.

In this paper, we characterize the scaling trends in VLA models across model architectures, hardware classes, and power budgets. We evaluate five representative VLAs, including three widely used baselines, VOTE~\cite{lin2025vote} (in three distinct configurations), and a newly developed QwenVLA model.  We profile their behavior on both state-of-the-art datacenter GPUs and resource-constrained edge devices under multiple power modes. Our study examines accuracy, memory footprint, latency, and throughput, revealing how model architectural choices (e.g., LLM backbone size, action head design, output tokenization) interact with hardware capabilities and power constraints. Beyond profiling, we distill actionable guidelines for selecting and optimizing VLA models tailored to diverse deployment scenarios, challenging common assumptions about edge–datacenter performance trade-offs. 

Paper organization: Section~\ref{sec:preliminaries_and_related_work} reviews relevant background and related work. Section~\ref{sec:experimental_setup} describes the hardware platforms, the VLA models evaluated, and our experimental methodology. Section~\ref{sec:results_analysis} presents analysis of accuracy, memory usage, latency, and throughput. Section~\ref{sec:conclusion} concludes with key takeaways and outlines directions for future work.











\section{Preliminaries and Related Work}
\label{sec:preliminaries_and_related_work}
Vision-Language-Action (VLA) models integrate visual perception and language understanding to directly generate control actions for robotic tasks~\cite{qu2025spatialvla, kim2025fine, kim2024openvla, zitkovich2023rt}. A representative example is OpenVLA~\cite{kim2024openvla}, a 7B-parameter open-source model trained on 970K robot demonstrations from the Open X-Embodiment (OXE) dataset~\cite{o2024open}, combining a Llama 2 language backbone with DINOv2 and SigLIP visual encoders. 
OpenVLA-OFT~\cite{kim2025fine} improves both inference efficiency and task success rate through parallel decoding, action chunking, continuous action representations, and an L1 regression objective function, achieving a 97.1\% success rate on the LIBERO benchmark~\cite{liu2023libero}.
SpatialVLA~\cite{qu2025spatialvla} extends the architecture of VLAs by introducing Ego3D position encoding and adaptive action grids for improved spatial reasoning across robots. Our prior work, \textbf{VOTE}~\cite{lin2025vote}, provides efficient and generalizable robotic manipulation. VOTE optimizes VLA architectures to generate fewer action tokens as compared to other action chunking methods, resulting in reduced inference latency and lower training costs. VOTE outperforms the state-of-the-art VLA models, achieving higher success rates and faster inference than OpenVLA, and proving effectiveness in real-world deployment scenarios.

While VLA models excel at integrating visual perception with language-driven action planning, little is known about how their performance scales across different model architectures and hardware classes, especially for fixed power budgets. Most prior work focuses on algorithmic advances or single-platform evaluations~\cite{kim2024openvla,qu2025spatialvla,kim2025fine}, without systematically analyzing how latency, throughput, and memory usage change across the edge-to-cloud spectrum. This lack of scaling insight limits our ability to make informed deployment decisions or design architectures tailored to specific hardware constraints.

VLA deployment environments vary significantly in their computational architecture. Edge platforms, such as the NVIDIA Jetson AGX Orin, integrate CPUs, GPUs and memory in a system-on-chip (SoC) design, optimized for power efficiency~\cite{nvidia_orin_manual}. The Orin supports multiple power modes (15W–50W), enabling trade-offs between performance and energy consumption~\cite{nvidia_orin_manual}. In contrast, cloud GPUs, such as the NVIDIA H100, are discrete accelerators with dedicated high-bandwidth memory, larger thermal envelopes, and significantly higher compute throughput~\cite{nvidia_h100_whitepaper_v1}. These architectural and power differences can lead to fundamentally different scaling behaviors for the same VLA workloads.

\section{Experimental Setup}
\label{sec:experimental_setup}

\begin{table}[t]
\centering
\caption{Jetson AGX Orin: Hardware specifications}
\vspace{-0.1cm}
\label{tab:orin_specification}
\resizebox{\linewidth}{!}{%
\begin{tabular}{ll}
\hline
\textbf{Feature}      & \multicolumn{1}{c}{\textbf{AGX Orin}}                        \\ \hline
\textbf{GPU} &
  \begin{tabular}[c]{@{}l@{}}\textbf{NVIDIA Ampere Architecture},\\ 2 GPCs, 8 TPCs, 16 SMs, 2048 CUDA cores (128/SM), 64 Tensor Cores,\\ 192KB L1 Cache/SM, 4MB L2 Cache, 1.3 GHz MAX Frequency\end{tabular} \\
\textbf{CPU} &
  \begin{tabular}[c]{@{}l@{}}12-core Arm Cortex-A78AE v8.2 (64-bit) in 3 clusters, 64KB L1i/L1d,\\ 3MB L2 (256KB/core), 6MB L3 (2MB/cluster), 4MB system cache,\\ MAX Frequency 2.2GHz\end{tabular} \\
\textbf{Memory}       & 32 GB 256-bit LPDDR5 @ 3200MHz, 204.8 GB/s                   \\
\textbf{Storage}      & 4TB NVMe M.2 SSD, Speeds Up to 7,450MB/s, and 32 GB eMMC 5.1 \\
\textbf{Power Budget} & up to 60 W (max configuration)                                     \\ \hline
\end{tabular}%
}
\vspace{-0.65cm}
\end{table}

\begin{table}[t]
\centering
\small
\caption{CPU/GPU configuration and maximum frequencies for different Jetson AGX Orin power modes. Lower power budgets reduce available cores and clock frequencies.}
\vspace{-0.1cm}
\label{tab:orin_power_modes}
\resizebox{0.8\linewidth}{!}{%
\begin{tabular}{@{}lcccc@{}}
\toprule
\textbf{Property}                   & \textbf{MAX}            & \textbf{50\,W} & \textbf{30\,W} & \textbf{15\,W} \\ \midrule
\textbf{Power budget (W)}           & n/a                      & 50           & 30           & 15           \\
\textbf{Online CPU cores}           & 12                       & 12           & 8            & 4            \\
\textbf{CPU max freq. (MHz)}        & 2201.6                   & 1497.6       & 1728.0       & 1113.6       \\
\textbf{GPU TPC count}              & 8                        & 8            & 4            & 3            \\
\textbf{GPU max freq. (MHz)}        & 1301.0                   & 828.75       & 624.75       & 420.75       \\
\textbf{Memory max freq. (MHz)}     & 3200                     & 3200         & 3200         & 2133         \\ \bottomrule
\end{tabular}%
}
\vspace{-0.1cm}
\end{table}

We evaluate representative VLA models on both edge and datacenter GPUs, measuring latency, throughput, and memory usage. The following subsections describe the hardware platforms and VLA models used in our experiments. 

\subsection{Hardware Platforms}
\noindent\textbf{Edge Computing Platform.} 
For evaluation on an edge device, we use the NVIDIA Jetson AGX Orin~\cite{nvidia_orin_manual}, a system-on-chip (SoC) platform designed for power-efficient, real-time AI workloads. The Orin integrates a CPU, a GPU and memory in a compact architecture.  Orin supports multiple configurable power modes (15\,W, 30\,W, 50\,W, and MAX), enabling us to explore trade-offs between performance and energy consumption. Table~\ref{tab:orin_specification} summarizes the key hardware specifications of Orin. The detailed CPU/GPU frequency scaling and core allocation for each power mode are shown in Table~\ref{tab:orin_power_modes}, illustrating how resource availability is progressively reduced under tighter power budgets. These features make the Orin representative of resource-constrained robotic platforms, where VLA models must operate within strict latency and power limits.

\begin{table}[t]
\centering
\caption{Key specifications of the datacenter-class GPUs used in our experiments, spanning multiple architecture generations and performance tiers.}
\vspace{-0.1cm}
\label{tab:server_class_specification}
\resizebox{\linewidth}{!}{%
\begin{tabular}{@{}lcccc@{}}
\toprule
\textbf{Feature}          & \textbf{H100} & \textbf{A100} & \textbf{A6000} & \textbf{V100} \\ \midrule
\textbf{Architecture}     & Hopper        & Ampere        & Ampere         & Volta         \\
\textbf{CUDA Cores / SMs} & 14,592 / 114  & 6,912 / 108   & 10,752 / 84    & 5,120 / 80    \\
\textbf{Tensor Cores}     & 456           & 432           & 336            & 640           \\
\textbf{L2 Cache}         & 50 MB         & 40 MB         & 6 MB           & 6 MB          \\
\textbf{Memory}           & 94 GB HBM3    & 40 GB HBM2e   & 48 GB GDDR6    & 32 GB HBM2    \\
\textbf{Memory Bandwidth} & 3.35 TB/s     & 1.6 TB/s      & 768 GB/s       & 900 GB/s      \\
\textbf{TDP}      & 700 W         & 400 W         & 300 W          & 300 W         \\ \bottomrule
\end{tabular}%
}
\vspace{-0.5cm}
\end{table}

\vspace{\noIndentSpace}

\noindent\textbf{Datacenter GPU Platforms.}
For datacenter-class evaluation, we use four discrete NVIDIA GPUs representing multiple architecture generations and performance tiers: H100 (Hopper)~\cite{nvidia_h100_whitepaper_v1}, A100 (Ampere)~\cite{nvidia_a100_architecture_whitepaper}, A6000 (Ampere)~\cite{nvidia_ga102_whitepaper_v2}, and V100 (Volta)~\cite{nvidia_volta_whitepaper_v1.1}. Unlike the integrated SoC design of AGX Orin, these GPUs feature dedicated high-bandwidth memory, large L2 caches, and significantly higher thermal design power (TDP), enabling substantially greater compute throughput. Table~\ref{tab:server_class_specification} summarizes their key specifications, including architecture, CUDA core count, memory type and capacity, and peak bandwidth. This diverse selection allows us to examine VLA performance scaling in high-power, high-throughput environments and contrast it with the constraints of edge deployments.


\subsection{Evaluated VLA Models}

We evaluate five Vision-Language-Action (VLA) models, including three established baselines (see Sec.~\ref{sec:preliminaries_and_related_work}): \textit{i)} OpenVLA~\cite{kim2024openvla}, \textit{ii)} SpatialVLA~\cite{qu2025spatialvla}, and \textit{iii)} OpenVLA-OFT~\cite{kim2025fine}. We also evaluate two VLA architectures developed by us: \textit{i)} VOTE~\cite{lin2025vote} and \textit{ii)} QwenVLA. Table~\ref{tab:model_summary} summarizes their key components, including language backbone, vision encoder, action head design, chunk size, and parameter count. 
\vspace{\noIndentSpace}

\noindent\textbf{VOTE}~\cite{lin2025vote} employs a Llama~2-7B backbone with DINOv2 and SigLIP encoders and a special token (ST)-based action head. By reducing the number of action tokens, VOTE minimizes latency without compromising accuracy. We evaluate three configurations to examine how output granularity influences scaling across hardware classes and power budgets.
\vspace{\noIndentSpace}

We propose \noindent\textbf{QwenVLA} to explore the impact of a smaller language backbone. It uses Qwen~2.5-1.5B~\cite{team2024qwen2} with the same DINOv2 and SigLIP vision stack and Cont-L1 action head as OpenVLA-OFT, maintaining a chunk size of 8. We adapt the Prismatic VLM architecture~\cite{karamcheti2024prismatic}, replacing its backbone with Qwen~2.5 and training on the LLaVA v1.5 data mixture~\cite{liu2024improved} before applying Optimized Fine-Tuning (OFT) on the LIBERO benchmark~\cite{liu2023libero}. To accommodate Qwen’s unique vocabulary and tokenization, we add special action tokens to its tokenizer. QwenVLA was fine-tuned using Optimized Fine-Tuning (OFT)~\cite{kim2025fine}, enabling it to achieve competitive performance with other models despite its smaller backbone.

Both VOTE and QwenVLA are fine-tuned on the LIBERO benchmark~\cite{liu2023libero} using AdamW with a learning rate of 1e-4 and 1e-3, respectively. Fine-tuning employs Low-Rank Adaptation (LoRA) with rank $r=32$ and $\alpha=16$, and a global batch size of 40 for VOTE and 64 for QwenVLA.

\begin{table}[t]
\caption{Summary of evaluated VLA models}
\vspace{-0.1cm}
\label{tab:model_summary}
\resizebox{\linewidth}{!}{%
\begin{threeparttable}
\begin{tabular}{@{}lccccc@{}}
\toprule
\textbf{Name} & \textbf{LLM} & \textbf{Vision} & \textbf{Action Head} & \textbf{Chunk} & \textbf{Parameters} \\ 
\midrule
OpenVLA     & LLaMA 2        & DINOv2+SigLIP   & DAT   & 1       & 7B     \\
SpatialVLA  & PaliGemma 2    & SigLIP+Ego3D    & AAG       & 4       & 4B     \\
OpenVLA-OFT & LLaMA 2        & DINOv2+SigLIP   & Cont-L1   & 8       & 7B     \\
QwenVLA     & Qwen 2.5       & DINOv2+SigLIP   & Cont-L1   & 8       & 2.6B   \\
VOTE        & LLaMA 2        & DINOv2+SigLIP   & ST        & 8, 16   & 7B     \\ 
\bottomrule
\end{tabular}
\begin{tablenotes}
\small
\item \textbf{DAT}: Discrete Action Tokens; \textbf{AAG}: Adaptive Action Grids; \textbf{Cont-L1}: continuous actions with L1 regression; \textbf{ST}: Special Token(s).
\end{tablenotes}
\end{threeparttable}%
}
\vspace{-0.5cm}
\end{table}

\subsection{Experimental Methodology}
\noindent\textbf{VLA Configurations.}
We use the standard implementations of OpenVLA, SpatialVLA, and OpenVLA-OFT. For VOTE, we evaluate three configurations: \textbf{VOTE-1T}, \textbf{VOTE-2T}, and \textbf{VOTE-MLP4}. VOTE-1T and VOTE-2T output one and two \texttt{<ACT>} tokens, respectively, with a chunk size of 8—yielding 8 actions for VOTE-1T and 16 actions for VOTE-2T—both using a 2-layer MLP action head. VOTE-MLP4 is a single \texttt{<ACT>} variant with a 4-layer MLP head, designed to improve performance when operating with a chunk size of 16. 
\vspace{\noIndentSpace}

\noindent\textbf{Benchmark.}
We evaluate VLA model accuracy using the LIBERO benchmark~\cite{liu2023libero}, which comprises a diverse set of robotic manipulation tasks in simulated environments. LIBERO has four task suites, evaluating the model’s understanding of spatial relationships (Spatial), object types (Object), task-oriented behaviors (Goal), and the model's ability to generalize to long-horizon tasks with diverse objects, layouts, and goals (Long).
\vspace{\noIndentSpace}

\noindent\textbf{Performance Evaluation Methodology.} 
We benchmark inference efficiency by comparing our models against baselines across all hardware platforms. The primary metrics are \emph{Latency} (i.e., the average time to generate an action chunk), and \emph{Throughput} (i.e., the number of actions generated per second). Each inference test processes a single 224$\times$224 RGB image and a fixed language prompt (“What action should the robot take to pick the cup?”). To ensure stable measurements, we perform several untimed warm-up runs before recording wall-clock times for 100 consecutive inferences, from which the average latency and throughput are computed.

\section{Results and Analysis}
\label{sec:results_analysis}
We analyze scaling trends in VLA performance across model architectures and hardware classes, while considering power constraints. Our evaluation covers three key dimensions: (1) task accuracy, measured on the LIBERO benchmark; (2) resource usage, focusing on peak memory usage; and (3) inference efficiency, characterized by latency and throughput across both edge and datacenter GPUs. We first compare model accuracies, then analyze memory usage, and finally present detailed latency and throughput results, highlighting the effects of Orin's power modes and the performance differences among four datacenter GPUs.
\vspace{\noIndentSpace}

\begin{table}[t]
\centering
\caption{Success rates (SR) on the LIBERO benchmark across four task suites: Spatial, Object, Goal, and Long.}
\vspace{-0.1cm}
\label{tab:models_accuracy_libero}
\resizebox{\linewidth}{!}{%
\begin{tabular}{@{}lccccc@{}}
\toprule
\textbf{Method} & \textbf{Spatial SR} & \textbf{Object SR} & \textbf{Goal SR} & \textbf{Long SR} & \textbf{Average} \\ \midrule
OpenVLA     & 84.7 & 88.4 & 79.2 & 53.7 & 76.5 \\
SpatialVLA  & 88.2 & 89.9 & 78.6 & 55.5 & 78.1 \\
OpenVLA-OFT & 96.2 & 98.3 & \textbf{96.2} & 90.7 & 95.3 \\
QwenVLA     & 77.8 & 90.0   & 82.8 & 64.6 & 78.8 \\
VOTE-1T     & \textbf{98.0} & \textbf{99.5} & 96.0 & \textbf{94.0} & \textbf{96.9} \\
VOTE-2T      & 96.0 & 98.5 & 94.0 & 91.0 & 94.9 \\
VOTE-MLP4    & 93.5 & 98.5 & 92.0 & 92.0 & 94.0 \\
\bottomrule
\end{tabular}%
}
\vspace{-0.5cm}
\end{table}

\begin{figure*}[ht]
    \centering
    \begin{subfigure}{\textwidth}
        \centering
        \includegraphics[width=\textwidth]{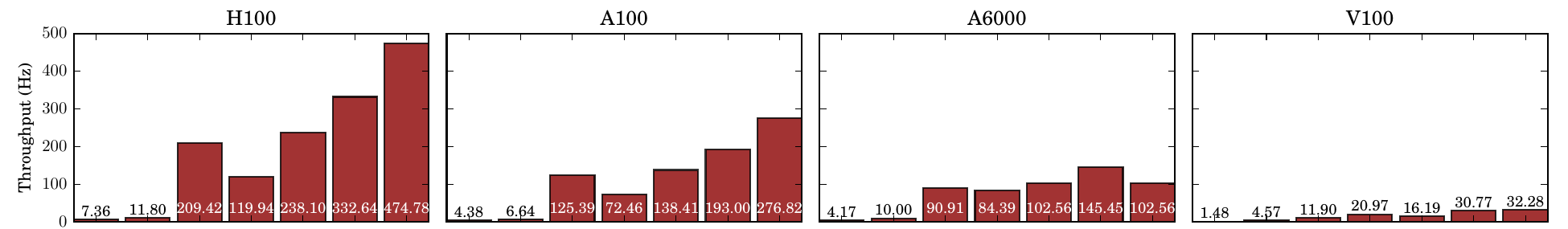}
        \caption{Datacenter GPU Throughput}
        \label{fig:throughput_hp}
    \end{subfigure}
    \vspace{1em} 
    \begin{subfigure}{\textwidth}
        \centering
        \includegraphics[width=\textwidth]{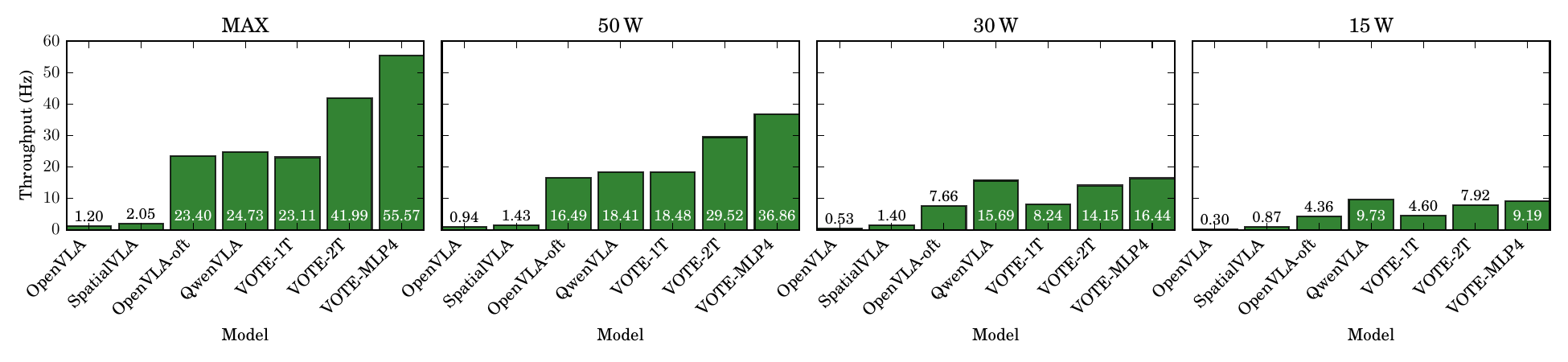}
        \vspace{-0.8cm}
        \caption{Orin Throughput by Power Mode}
        \label{fig:throughput_orin}
    \end{subfigure}    
    \vspace{-0.7cm}
    \hrule
    \vspace{-0.2cm}
    \caption{\textit{Throughput (Hz) for each evaluated VLA model across (a) four datacenter GPUs and (b) Jetson AGX Orin under different power modes. Results highlight scaling trends with hardware class, power budget, and model architecture.}}
    \label{fig:throughput_comparison}
    \vspace{-0.5cm}
\end{figure*}

\textbf{VLA Model Accuracy Comparison.} 
Each model is evaluated on the four LIBERO task suites, each containing 10 tasks with 20 repetitions, for a total of 200 trials per suite. Table~\ref{tab:models_accuracy_libero} shows that our VOTE variants achieve the highest average success rates (SR) across all suites. \textbf{VOTE-1T} attains the top overall performance at 96.9\%, outperforming the strongest baseline, OpenVLA-OFT (95.3\%), and achieves the highest SR in three out of four suites. \textbf{VOTE-2T} and \textbf{VOTE-MLP4} trade a small drop in accuracy (up to 2.9\%) for higher throughput, which is enabled by larger output chunks, a trade-off explored further in Fig.~\ref{fig:throughput_comparison}. \textbf{QwenVLA}, despite its much smaller 1.5B backbone, surpasses the larger OpenVLA baseline in average SR (78.8\% vs.\ 76.5\%) and is particularly effective for the LIBERO-Object and LIBERO-Goal suites, demonstrating that competitive task performance can be achieved with reduced model size.
\vspace{\noIndentSpace}

\begin{figure}[t]
    \centering
    \includegraphics[width=\linewidth]{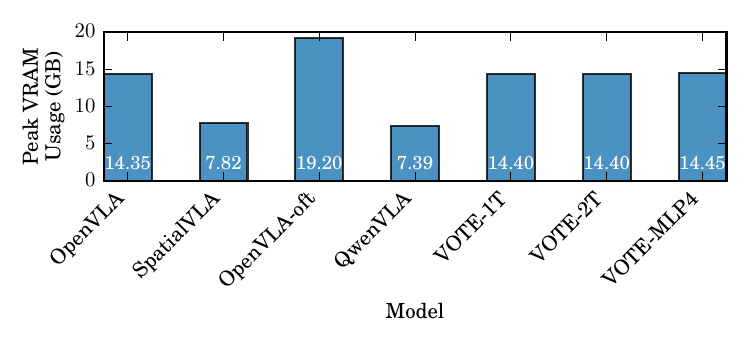}
    \vspace{-0.6cm}
    \hrule
    \vspace{0.1cm}
    \caption{\textit{Peak VRAM usage for each evaluated VLA model during inference on the NVIDIA Jetson AGX Orin.}}
    \label{fig:memory_usage}
    \vspace{-0.7cm}
\end{figure}

\noindent\textbf{Peak Memory Usage.} 
Fig.~\ref{fig:memory_usage} shows the peak VRAM usage for each model during inference. Among the baselines, OpenVLA and OpenVLA-OFT are the most memory-intensive, requiring 14.35\,GB and 19.20\,GB, respectively, while SpatialVLA is significantly less demanding at 7.82\,GB. QwenVLA has the lowest usage overall (7.39\,GB), reflecting its smaller 1.5B backbone. All VOTE configurations have similar memory footprints (14.40–14.45\,GB), comparable to OpenVLA, despite architectural differences. These results indicate that memory usage is primarily driven by backbone size and vision encoder choice, with action head variations (e.g., token count or MLP depth) having negligible impact.
\vspace{\noIndentSpace}

\begin{figure}[t]
    \centering
    \begin{subfigure}{\linewidth}
        \centering
        \includegraphics[width=\linewidth]{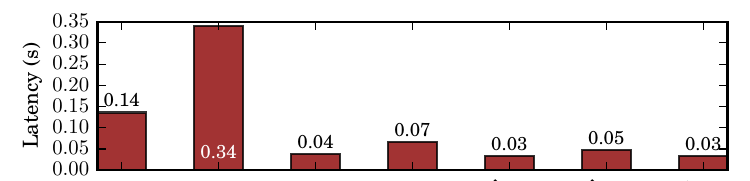}
        \caption{H100 Latency}
        \vspace{-0.0cm}
        \label{fig:latency_h100}
    \end{subfigure}
    \vspace{1em} 
    \begin{subfigure}{\linewidth}
        \centering
        \includegraphics[width=\linewidth]{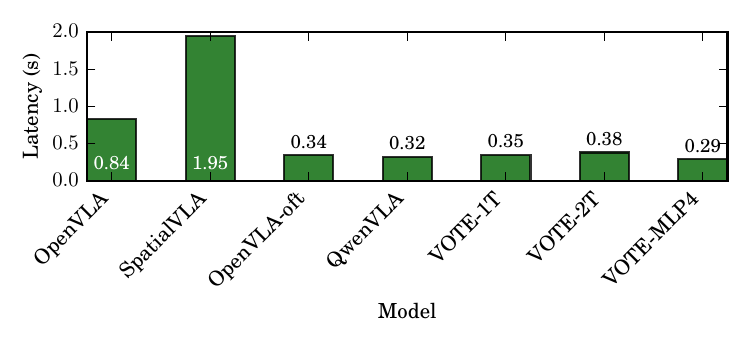}
        \vspace{-0.9cm}
        \caption{AGX Orin Latency (MAX) }
        \label{fig:latency_orin}
    \end{subfigure}
    \vspace{-0.8cm}
    \hrule
    \vspace{-0.1cm}
    \caption{\textit{Per-chunk latency for each VLA model evaluated on the H100 datacenter GPU and Jetson AGX Orin (MAX power mode). The H100 achieves latencies roughly an order of magnitude smaller than the Orin across all models. VOTE configurations are consistently competitive on both platforms, with VOTE-MLP4 achieving the lowest latency on Orin.}}
    \label{fig:latency_comparison}
    \vspace{-0.7cm}
\end{figure}

\textbf{Latency Comparison (Orin MAX vs. H100).} 
Fig.~\ref{fig:latency_comparison} compares per-chunk latency on the highest-performance datacenter GPU (H100) and the NVIDIA Jetson AGX Orin in MAX power mode. As expected, the H100 achieves substantially lower latencies across all models, with values ranging from 0.03\,ms (VOTE-1T, VOTE-MLP4) to 0.34\,ms (SpatialVLA). On Orin, latencies increase by roughly an order of magnitude, ranging from 0.29\,ms (VOTE-MLP4) to 1.95\,ms (SpatialVLA). Notably, VOTE configurations maintain competitive latencies on both platforms, with VOTE-MLP4 achieving the lowest latency overall on Orin. These results establish the performance gap between edge and datacenter GPUs, providing context for the throughput scaling trends analysis.
\vspace{\noIndentSpace}

\noindent\textbf{Throughput Scaling Across Datacenter GPUs.}
Fig.~\ref{fig:throughput_hp} shows throughput across four datacenter GPUs. The H100 consistently delivers the highest performance, with VOTE-MLP4 reaching 474.78\,Hz---over 64$\times$ faster than OpenVLA on the same hardware. The A100 follows a similar scaling pattern, albeit with lower absolute values, with VOTE-MLP4 sustaining 276.82\,Hz. The A6000 maintains strong performance for smaller models such as SpatialVLA (10.00\,Hz) and QwenVLA (84.39\,Hz), but these advantages diminish for larger, more compute-intensive models. The V100, constrained by older architecture and lower memory bandwidth, exhibits the lowest throughput overall, with VOTE-MLP4 peaking at 32.28\,Hz. Across all datacenter GPUs, throughput improvements from VOTE configurations scale with output chunk size and MLP depth, while smaller-backbone models such as QwenVLA offer competitive performance per parameter, but do not match the absolute throughput of optimized VOTE variants.
\vspace{\noIndentSpace}

\noindent\textbf{Throughput Scaling Across Edge Power Budgets.}
Fig.~\ref{fig:throughput_orin} illustrates how throughput scales strongly with power constraints on AGX Orin, highlighting the impact of power budget on sustained inference rates. In MAX mode, VOTE-MLP4 reaches 55.57\,Hz, more than 46$\times$ faster than OpenVLA (1.20\,Hz) under the same conditions. Even at reduced power levels, the relative ordering of models remains consistent—VOTE configurations lead, followed by QwenVLA and OpenVLA-OFT, with OpenVLA and SpatialVLA following. Throughput reductions are non-linear with power scaling; dropping from 50\,W to 30\,W, which results in sharper performance losses, particularly for compute-heavy models, while smaller models like QwenVLA retain a higher fraction of their MAX-mode throughput. These results suggest that both architecture choice and power allocation are critical levers for balancing efficiency and responsiveness in edge deployments.
\vspace{\noIndentSpace}

\noindent\textbf{Throughput Scaling Across Edge and Datacenter.}  
Comparing Orin in MAX mode to the fastest datacenter GPU (H100) underscores the edge–cloud performance gap: VOTE-MLP4 runs 8.5$\times$ faster on H100, and even smaller models like QwenVLA are roughly 4.8$\times$ faster in the cloud. This disparity widens for models with larger backbones or more complex decoders, where datacenter GPUs benefit disproportionately from higher memory bandwidth and greater parallelism. However, the relative scaling trends remain similar across hardware classes—architectures optimized for chunked decoding (e.g., VOTE-2T, VOTE-MLP4) consistently lead, followed by efficient smaller-backbone models like QwenVLA. Notably, Orin in MAX mode with VOTE-MLP4 achieves 55.57\,Hz, surpassing the throughput of the V100 datacenter GPU (32.28\,Hz). This demonstrates that modern high-end edge devices can outperform older datacenter hardware, challenging the assumption that any datacenter GPU will necessarily exceed edge performance.

\section{Conclusion and Future Work}
\label{sec:conclusion}
In this work we characterized scaling trends in Vision-Language-Action (VLA) models across model architectures, hardware classes, and for limited power budgets. Through a systematic evaluation spanning edge (Jetson AGX Orin in multiple power modes) and datacenter GPUs (H100, A100, A6000, V100), we analyzed accuracy, memory usage, latency, and throughput for five representative VLA architectures, including two of our own~\cite{lin2025vote}.

Our results reveal that architectures optimized for chunked decoding, such as VOTE-2T and VOTE-MLP4, deliver the highest throughput across hardware classes, with only minor accuracy trade-offs relative to VOTE-1T. Model memory usage is primarily dictated by backbone size and vision encoder choice, with smaller-backbone designs such as QwenVLA achieving the lowest footprint while maintaining competitive accuracy. Latency and throughput scale predictably with available compute, but modern high-end edge devices, such as Orin in MAX mode, can surpass older datacenter GPUs.

These findings provide actionable guidance for selecting and configuring VLA models based on deployment constraints and performance priorities. Future work will extend this analysis to additional model architectures, quantization strategies, and real-world robotic deployments to further optimize VLA inference under practical constraints.

\balance
\bibliographystyle{./IEEEtrans}
\bibliography{./annot.bib}

\end{document}